\documentclass[sigconf]{acmart}

\usepackage{bm}
\usepackage{multirow}
\usepackage{booktabs}

\newcommand{\shortname}{MTM}
\newcommand{\longname}{\textbf{m}ulti-scale \textbf{t}oken \textbf{m}ixing transformer}

\newcommand{\nan}{\texttt{NaN}}
\newcommand{\cls}{\texttt{[CLS]}}

\copyrightyear{2025}
\acmYear{2025}
\setcopyright{acmlicensed}
\acmConference[KDD '25]{Proceedings of the 31st ACM SIGKDD Conference on Knowledge Discovery and Data Mining V.2}{August 3--7, 2025}{Toronto, ON, Canada}
\acmBooktitle{Proceedings of the 31st ACM SIGKDD Conference on Knowledge Discovery and Data Mining V.2 (KDD '25), August 3--7, 2025, Toronto, ON, Canada}
\acmDOI{10.1145/3711896.3737058}
\acmISBN{979-8-4007-1454-2/2025/08}
\begin{document}

\title[\shortname{}]{\shortname{}: A Multi-Scale Token Mixing Transformer for \\ Irregular Multivariate Time Series Classification}

\author{Shuhan Zhong}
\orcid{0000-0003-4037-4288}
\affiliation{
  \institution{The Hong Kong University of Science and Technology}
  \city{Hong Kong SAR}
  \country{China}}
\email{szhongaj@cse.ust.hk}

\author{Weipeng Zhuo}
\authornote{Corresponding Author}
\orcid{0000-0002-1810-7071}
\affiliation{
  \institution{Beijing Normal-Hong Kong Baptist University}
  \city{Zhuhai}
  \country{China}}
\email{weipengzhuo@uic.edu.cn}

\author{Sizhe Song}
\orcid{0000-0001-8344-830X}
\affiliation{
  \institution{The Hong Kong University of Science and Technology}
  \city{Hong Kong SAR}
  \country{China}}
\email{ssongad@cse.ust.hk}

\author{Guanyao Li}
\orcid{0000-0002-3950-9360}
\affiliation{
  \institution{Guangzhou Urban Planning and Design Survey Research Institute}
  \city{Guangzhou}
  \country{China}}
\email{gliaw@connect.ust.hk}

\author{Zhongyi Yu}
\orcid{0000-0003-2859-7008}
\affiliation{
  \institution{Beijing Normal-Hong Kong Baptist University}
  \city{Zhuhai}
  \country{China}}
\email{zhongyiyu@uic.edu.cn}

\author{S.-H. Gary Chan}
\orcid{0000-0003-4207-764X}
\affiliation{
  \institution{The Hong Kong University of Science and Technology}
  \city{Hong Kong SAR}
  \country{China}}
\email{gchan@cse.ust.hk}

\renewcommand{\shortauthors}{Shuhan Zhong et al.}

\begin{abstract}
  Irregular multivariate time series (IMTS) is characterized by the lack of synchronized observations across its different channels. In this paper, we point out that this channel-wise asynchrony can lead to poor channel-wise modeling of existing deep learning methods. To overcome this limitation, we propose \shortname{}, a \longname{} for the classification of IMTS. We find that the channel-wise asynchrony can be alleviated by down-sampling the time series to coarser timescales, and propose to incorporate a masked concat pooling in \shortname{} that gradually down-samples IMTS to enhance the channel-wise attention modules. Meanwhile, we propose a novel channel-wise token mixing mechanism which proactively chooses important tokens from one channel and mixes them with other channels, to further boost the channel-wise learning of our model. Through extensive experiments on real-world datasets and comparison with state-of-the-art methods, we demonstrate that \shortname{} consistently achieves the best performance on all the benchmarks, with improvements of up to $3.8\%$ in AUPRC for classification.
\end{abstract}





\maketitle

\newcommand\kddavailabilityurl{https://doi.org/10.5281/zenodo.15600519}

\ifdefempty{\kddavailabilityurl}{}{
  \begingroup\small\noindent\raggedright\textbf{KDD Availability Link:}\\
  The source code of this paper has been made publicly available at \url{\kddavailabilityurl}.
  \endgroup
}

\section{Introduction}

\begin{figure}[tp]
  \centering
  \includegraphics[width=\linewidth]{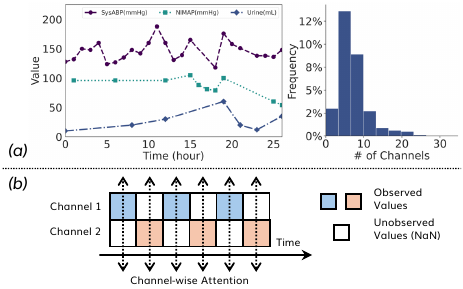}
  \caption{(a) Left: An example of irregular multivariate time series. Right: The distribution of number of channels observed per timepoint in the P12 dataset. (b) A failure example of channel-wise attention in existing Transformers. Channels 1 and 2 have no synchronized observation, and will be processed as independent even though they are potentially correlated.}
  \label{fig:asyn}
\end{figure}

Irregular multivariate time series (IMTS) is a common data modality in various science domains and real-world applications \cite{jensen2012mining,schulz1997spectrum,ruf1999lomb,scargle1982studies}. An IMTS is a sequence of \emph{irregularly sampled} observations indexed in time order, where observations of different variates form the ``channel'' dimension. The irregularity of sampling is characterized by the uneven time interval between consecutive observations, as well as partially observed channels. Specifically, we find that the irregularity of IMTS can lead to a serious lack of synchronized observations across its different channels, as shown in Figure \ref{fig:asyn}(a). Such channel-wise asynchrony presents significant challenges for deep learning methods to model channel-wise correlation of IMTS data. Early efforts on IMTS analysis mainly focus on improving the temporal modeling of IMTS data by considering the uneven time interval with specially designed deep learning modules. However, most of them have not fully considered the channel-wise asynchrony \cite{weerakody2021review,chen2018neural,tipirneni2022self}. They either bypass channel-wise asynchrony by simply processing different channels independently, or impute missing channels to obtain a fixed-size vector per timepoint as model input, which leads to poor channel-wise modeling of the data.

In light of this, some methods propose Transformer models that tokenize each single observation in an IMTS and rely on attention modules along the temporal and channel dimensions to model their correlations \cite{zhang2023warpformer,yalavarthi2024grafiti,wei2023compatible}. Nevertheless, we argue that the channel-wise attention in these methods is also \emph{ineffective} for channel-wise modeling, as it relies on the \emph{synchronized} observations. Due to the channel-wise asynchrony, such channel-wise attention is always computed over a small part of channels. Moreover, if some channels do not have any synchronized observations, which is a common case in real-world IMTS data, no channel-wise attention will be computed between these channels (Figure \ref{fig:asyn}(b)). These channels are thus processed independently despite that the potential correlations between them may help the downstream task. As presented in Table \ref{tab:tf}, we find almost no performance drop when substituting the channel-wise attention module with simple multi-layer perceptrons (MLPs) in such Transformers.

In this work, we overcome the limitation on channel-wise modeling of existing methods in two aspects, i.e., to improve the effectiveness of the channel-wise attention with multi-scale down-sampling, and to propose a new token mixing mechanism to further enhance the channel-wise modeling. First, we find that the lack of synchronized observations can be gradually alleviated when we down-sample the IMTS to coarser timescales. However, the limited existing works on this are either uni-scale patching \cite{zhang2024irregular}, or cannot guarantee to mitigate the channel-wise asynchrony \cite{zhang2023warpformer}. On the other hand, different from the general down-sampling on regular dense data, the irregularity and sparsity of IMTS makes it more critical to retain information during the down-sampling process. The effectiveness of general down-sampling methods remains \emph{unknown} on IMTS. In this work, we conduct a comprehensive comparison of potential down-sampling methods, and find that a simple concatenation of max and average temporal pooling, with the missing values masked, is the most effective IMTS down-sampler for our purpose.

Furthermore, although down-sampling can help cross-channel feature learning in coarser timescales, it sacrifices the temporal resolution of the data and still suffers in finer timescales. To solve this problem, we propose a novel token mixing mechanism to further enhance the channel-wise feature learning without down-sampling. The intuition behind is to mix tokens across channels \emph{even if they are not synchronized}, which is achieved by proactively choosing important tokens from one channel, and mixing them with unsynchronized tokens in other channels. Specifically, based on the attention scores from a set of specially designed \cls{} tokens, we find out the importance of tokens within each channel. We then choose the pivotal tokens to fill the missing channels of the same timepoint. Then, we mix the information of each channel's original tokens with the pivotal tokens from other channels with another self-attention step. After the mixing operation, we reset the missing channels to keep the original sampling patterns \emph{unchanged}. Through this token mixing mechanism, the pivotal features from one channel can be effectively mixed with other channels in both coarse and fine timescales, such that channel-wise features can be better captured.

\begin{table}[tp]
  \centering
  \small
  \caption{The channel-wise attention module in Transformer is ineffective for IMTS classification.}
  \begin{tabular}{c|c|c|c}
    \toprule
    Transformer           & P12 AUROC                  & P19 AUROC                  & PAM F1-Score               \\
    \midrule
    w/ Channel Attention  & 86.1 \scriptsize{$\pm$1.6} & 87.3 \scriptsize{$\pm$3.2} & 95.8 \scriptsize{$\pm$1.2} \\
    w/o Channel Attention & 85.8 \scriptsize{$\pm$1.6} & 87.6 \scriptsize{$\pm$2.7} & 96.1 \scriptsize{$\pm$0.7} \\
    \bottomrule
  \end{tabular}%
  \label{tab:tf}%
\end{table}%

To this end, we propose \shortname{}, a \longname{} for IMTS classification that effectively integrates the masked concat pooling and the token mixing mechanism. We carry out extensive experiments on 3 real-world datasets with various experiment settings, and compare \shortname{} with state-of-the-art methods. The results show that \shortname{} consistently achieves the best performance on all the benchmarks, with significant improvements of up to 3.8\% in classification AUPRC.

To summarize, we make the following contributions:
\begin{itemize}
  \item \emph{A thorough investigation} into the channel-wise asynchrony in IMTS and how it impacts channel-wise modeling of existing deep learning models for IMTS.
  \item \emph{A masked concat pooling method} that down-samples IMTS to mitigate its channel-wise asynchrony and improves the effectiveness of channel-wise attention, which is designed based on a comprehensive comparison of different down-sampling methods on IMTS.
  \item \emph{A channel-wise token mixing mechanism} to further enhance the channel-wise feature learning of IMTS in all timescales.
  \item \emph{A multi-scale token mixing Transformer model \shortname{}} that integrates the pooling and token mixing modules for the classification of IMTS, with extensive experiments to validate its effectiveness.
\end{itemize}

The remainder of this paper is organized as follows: We first review related works in Section~\ref{sec:rw}, and formally define the problem we study in Section~\ref{sec:probdef}. Afterwards, we elaborate on the design of \shortname{} and its modules in Section~\ref{sec:mtm}. We discuss the experimental results in Section~\ref{sec:exp} and conclude in Section~\ref{sec:conclu}.

\section{Related Works}
In this section, we review related works below from two aspects. First, we summarize and compare prior studies on IMTS thoroughly in Section~\ref{sec:rel1}. Second, although token down-sampling has not been applied for the analysis of IMTS before, we also list relevant works analyzing regular data in the literature, and discuss the down-sampling for IMTS in \ref{sec:rel2}.
\label{sec:rw}
\subsection{Modeling Irregular Multivariate Time Series}
\label{sec:rel1}
IMTS is prevalent in various real-world domains such as healthcare \cite{jensen2012mining}, climate science \cite{schulz1997spectrum}, traffic \cite{xu2025mmpath}, biology \cite{ruf1999lomb}, and astronomy \cite{scargle1982studies}. Despite the recent success of deep learning in the analytics of regular MTS \cite{zhong2024multi,qiu2024tfb,qiu2025duet,nie2023time,zeng2022transformers}, how to tackle the irregularities in IMTS with deep learning still remains a challenging problem \cite{shukla2021survey}. Earliest approaches \cite{marlin2012unsupervised,li2020learning,lipton2016directly,shan2023nrtsi,xu2021mufasa,zhang2021learning,xu2018raim,ma2020adacare} basically rely on data imputation to preprocess the time series to be regular and leverage general time series models. Nevertheless, this will inevitably introduce bias, artifacts, or cause information loss to the dataset, thus impacting the analysis of the data \cite{shukla2021survey,zhang2022graph}.

To address this problem, efforts have been made to directly model IMTS without imputation. Recent works try to adapt deep sequence models to deal with the temporal irregularity of IMTS, such as introducing the time interval information into the state transition of Recurrent Neural Networks \cite{che2018recurrent,weerakody2021review,neil2016phased}, learning neural ordinary differential equations along the temporal dimension \cite{chen2018neural,rubanova2019latent,chen2024contiformer,de2019gru,schirmer2022modeling,bilovs2021neural,scholz2023latent}, and encoding the irregular time for attention mechanism \cite{luo2020hitanet,tipirneni2022self}. However, most of these methods still cannot handle channel-wise asynchrony. In light of this, some Transformer-based methods \cite{zhang2023warpformer,wei2023compatible,yalavarthi2024grafiti} propose to tokenize each single observation in an IMTS by encoding the corresponding time and channel information together with the observed value, and use attention modules to capture their pair-wise correlations. Since a global attention over all tokens is prohibitive considering the quadratic complexity of self-attention, these models generally compute the attention along the temporal and channel dimensions separately. However, the lack of synchronized observations in IMTS limits the ability of channel-wise attention to learn channel-wise correlations, which leads to poor channel-wise modeling of these methods.

There are only few discussion on the channel-wise modeling problem of IMTS. t-PatchGNN \cite{zhang2024irregular} proposes a patching approach to transform the IMTS into uniform and aligned temporal patches to avoid the irregularity. Despite it mitigates the asynchrony to some extent, the patching is applied only once at the beginning of their model, where multi-scale down-sampling is needed for better channel-wise modeling. Warpformer \cite{zhang2023warpformer} is the first multi-scale model for IMTS, with a learnable irregular warping approach to project IMTS into multiple timescales. However, this approach does not always help align different channels. In some cases, it even exacerbates the asynchrony by projecting originally aligned observations apart. ViTST \cite{li2023time} converts IMTS into images and utilizes Vision Transformer with multi-scale modeling abilities for classification. Nonetheless, the conversion is complicated and time-consuming. Moreover, the computation overhead is much larger than other methods. In summary, all these methods are designed to improve the effectiveness of channel-wise learning on coarser timescales. Besides, they all rely on imputed data to some extent, which hampers their modeling abilities.

In contrast, our \shortname{} addresses the channel-wise asynchrony in two aspects, i.e., improving the existing channel-wise learning module by gradually down-sampling the data to coarser timescales, and introducing a new channel-wise feature learning mechanism to enhance channel-wise modeling in all timescales. As shall be demonstrated in Section~\ref{sec:exp}, our proposed \shortname{} shows significantly better performance than existing methods, without the need of any data imputation.

\begin{figure*}[tbp]
  \centering
  \includegraphics[width=\linewidth]{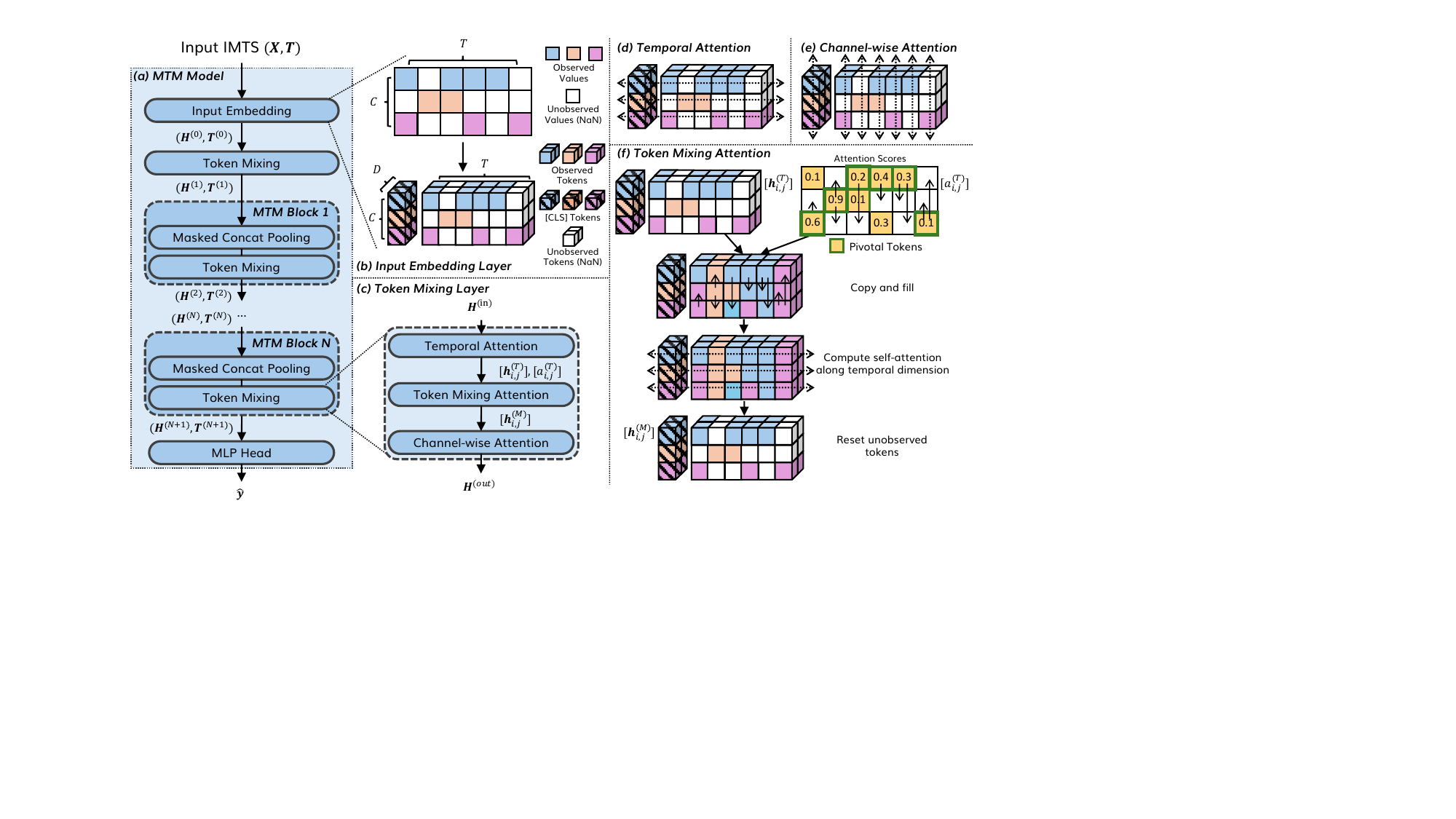}
  \caption{Overview of \shortname{}.}
  \label{fig:ov}
\end{figure*}

\subsection{Token Down-Sampling in Transformers}
\label{sec:rel2}
Token down-sampling can be broadly categorized into the structured down-sampling and the unstructured down-sampling based on how they partition and reduce the target dimension. Structured down-sampling methods, such as max pooling and average pooling, partition the dimension into fixed-sized patches, with reduction operation performed within each patch. On the contrary, unstructured down-sampling methods offer a more flexible approach by adapting the partition \cite{haurum2023tokens}. Recent methods \cite{liang2022patches,bolya2022token,marin2023token,rao2021dynamicvit} generally rely on a scoring function, such as the attention scores in the Transformer, to partition the dimension into varying sized patches, which allows the model to balance its focus on regions of interest and potentially leading to improved performance.

Since these methods are mostly studied for dense and regular data modalities such as regular time series and images, they focus more on extracting essential features from a redundancy of data and reducing computational cost. However, the irregularity and sparsity of IMTS make it more critical to retain information during the down-sampling process compared with regular data. In addition, as channel-wise asynchrony has not been attended before, the effectiveness of general down-sampling methods to mitigate the channel-wise asynchrony remains unknown on IMTS. In this work, we identify the value of down-sampling in tackling the channel-wise asynchrony, and conduct a comprehensive comparison of potential down-sampling methods to find that a simple concatenation of max and average temporal pooling, with the missing values masked, is the most effective IMTS down-sampler for our purpose.

\section{Problem Definition}
\label{sec:probdef}
An IMTS of $T$ timepoints and $C$ channels is a matrix of chronologically ordered observations $\bm{X}=[x_{i,j}]\in\left(\mathbb{R} \cup \{\nan\}\right)^{T\times C}$ together with its observation time $\bm{T}=[t_i]$, where $x_{i,j}$ is observation of channel $j$ at time $t_i$ for $i\in[1,T]$ and $j\in[1,C]$. Specifically, the \emph{irregularity} of IMTS refers to that $t_{i+1}-t_{i}$ may not be constant, and each observation may contain missing channels which makes $x_{i,j}=\nan$. In this paper, we consider the series-level classification of IMTS that predicts one categorical label for each time series. Given a dataset $\mathcal{D}$ containing sample pairs $(\bm{X},\bm{T},y)$, where $\bm{X}$ and $\bm{T}$ are the input IMTS and its observation time, $y\in\{0,\ldots,M-1\}$ is the categorical label of $M$ classes, the classification problem is to obtain an optimal function $\mathcal{F}(\cdot)$ that maps the input $(\bm{X},\bm{T})$ to its corresponding label as $\hat{y}=\mathcal{F}(\bm{X}, \bm{T})$ such that the difference between $\hat{y}$ and the ground truth $y$ is minimized.

\section{\shortname{}}
\label{sec:mtm}
In this section, we first overview the architecture and workflow of our proposed \shortname{} in Section~\ref{sec:ov}, and then elaborate on the designs of \shortname{} from Sections \ref{sec:init} to \ref{sec:tm}.

\subsection{\shortname{} Overview}

\label{sec:ov}
Figure \ref{fig:ov}(a) shows the overall architecture of \shortname{}. \shortname{} comprises an Input Embedding layer, a Token Mixing layer, and a stack of $N$ \shortname{} Blocks. Each \shortname{} Block is further composed of a Masked Concat Pooling layer and a Token Mixing layer. In \shortname{}, the input IMTS $\bm{X}$ is first embedded by the Input Embedding layer to obtain the high dimensional token embeddings of the observations. Meanwhile, in the Input Embedding layer, we also introduce a set of special \cls{} tokens attached together with the observation tokens. These \cls{} tokens will be used in our token mixing mechanism in the Token Mixing layers, and also be used to predict the final class label at the end. We denote the initial token embeddings as $\bm{H}^{(0)}$. After the Input Embedding layer, $\bm{H}^{(0)}$ is first fed into a Token Mixing layer with the output denoted as $\bm{H}^{(1)}$, followed by a stack of $N$ \shortname{} Blocks for feature extraction. Denote $\bm{T}^{(1)}=\bm{T}$, then the $n$-th \shortname{} Block of \shortname{} takes the token embeddings and observation time $(\bm{H}^{(n)},\bm{T}^{(n)})$ as input, down-samples the data and mixes the feature to get $(\bm{H}^{(n+1)},\bm{T}^{(n+1)})$. The output of the last \shortname{} Block in \shortname{} is $(\bm{H}^{(N+1)},\bm{T}^{(N+1)})$. We extract the embeddings of the \cls{} tokens from $\bm{H}^{(N+1)}$ and compute the max pooling of these tokens, followed by an MLP layer to get the class label prediction $\hat{y}$.

\subsection{Input Embedding}
\label{sec:init}
\shortname{} treats each single observation in an IMTS as a token, and employs the attention mechanism to extract features from the tokens for classification. In the Input Embedding layer of \shortname{} shown as Figure \ref{fig:ov}(b), we map each observation in the input IMTS $\bm{X}$ to a high dimensional token embedding $\bm{Z}=[\bm{z}_{i,j}]\in\left(\mathbb{R} \cup \{\nan\}\right)^{T\times C\times D}$ by considering the observed value, observation time, and channel of each observation, with
\begin{equation}
  \label{eq:embed}
  \begin{aligned}
    \bm{z}_{i,j}=x_{i,j}\cdot \bm{e}_{j}+\textrm{PE}(t_i), (i\in [1,T],j\in[1,C],x_{i,j}\ne\nan{}),
  \end{aligned}
\end{equation}
where $\bm{e}_j\in\mathbb{R}^D$ is the trainable embedding vector for the $j$-th channel, $D$ is the size of the embedding dimension, and $\textrm{PE}(\cdot)$ is the sinusoidal positional encoding function of $D$ dimensions. To facilitate the computation in the Token Mixing layers, we introduce a tailor-made \cls{} token for each of the $C$ channels. The embeddings of the \cls{} tokens are randomly initialized as $D$ dimensions and trainable. We denote the \cls{} tokens as $\bm{S}=[\bm{s}_j]\in\mathbb{R}^{C\times D}$. We concatenate the \cls{} tokens together with the observation tokens to obtain
\begin{equation}
  \label{eq:cls}
  \begin{aligned}
    \bm{H}^{(0)}=[\bm{S}||\bm{Z}]=[\bm{h}^{(0)}_{i,j}],(i\in [0,T],j\in[1,C]),
  \end{aligned}
\end{equation}
where $\bm{h}^{(0)}_{0,j}$ denotes the \cls{} token for channel $j$.

\subsection{Masked Concat Pooling Layer}
\label{sec:mcp}

\begin{figure}[tbp]
  \centering
  \includegraphics[width=\linewidth]{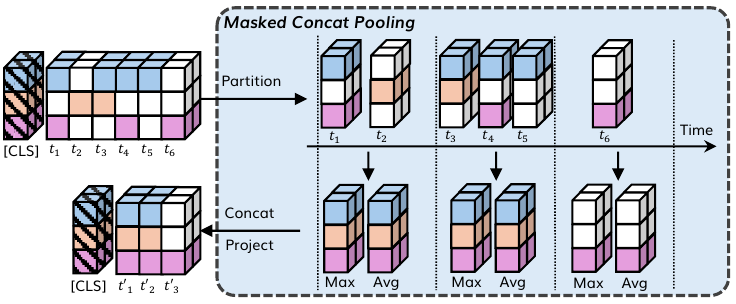}
  \caption{An example of IMTS down-sampling by masked concat pooling.}
  \label{fig:ds}
\end{figure}

To mitigate the asynchrony between channels of IMTS, in \shortname{} we propose to down-sample the data along the temporal dimension by merging temporally adjacent observations, such that tokens can be better aligned along the channel dimension for the attention mechanism to effectively learn the channel-wise features. In \shortname{}, we employ the Masked Concat Pooling layers to perform the down-sampling as shown in Figure \ref{fig:ds}. Each Masked Concat Pooling layer is associated with a down-sampling rate $R$. In each layer we first partition the original time span into non-overlapping patches of size $R$, then compute the max and average of token embeddings within each patch. Timepoints with no observations are masked out and will not be computed. We then concatenate the max and average token embeddings of each patch, and use a trainable linear projection to map the embeddings of size $2D$ back to $D$. The projected embeddings for the corresponding timepoints are used in the down-sampled timescale. Last, we update the observation time of the tokens accordingly to match the down-sampled timescale, and fed the obtained token embeddings and their observation time to next layers for further processing.

\subsection{Token Mixing Layer}
\label{sec:tm}
The channel-wise asynchrony of IMTS makes the channel-wise attention in existing Transformer models less effective, especially without down-sampling. To this end, we propose Token Mixing layer in \shortname{} by introducing a novel token mixing mechanism that share important information of one channel across other channels, as it is more likely to be important for other channels.

As is shown in Figure \ref{fig:ov}(c), a Token Mixing layer is composed of a Temporal Attention module, a Token Mixing Attention module, and a Channel-wise Attention module. Here we omit the layer and block index in the notations for simplicity, and denote the input token embeddings as $\bm{H}^{(in)}=[\bm{h}_{i,j}^{(in)}],(i\in [0,T],j\in[1,C])$, where $[\bm{h}_{0,:}^{(in)}]$ are embeddings of the \cls{} tokens, and $[\bm{h}_{1:,:}^{(in)}]$ are embeddings of the observation tokens. The Temporal Attention module takes $\bm{H}^{(in)}$ as input, and compute the temporal attention by treating tokens from each channel as a sequence (Figure \ref{fig:ov}(d)).
Taking the $j$-th channel as an example, we compute the pairwise attention scores $[a_{i,k}^{(T)}]$ of sequence $[\bm{h}_{0,j}^{(in)},\ldots,\bm{h}_{T,j}^{(in)}]$, and update the token embeddings as
\begin{equation}
  \label{eq:tattn}
  \begin{aligned}
     & a_{i,k}^{(T)}=\frac{(\bm{W}_Q^{(T)}\bm{h}_{i,j}^{(in)})(\bm{W}_K^{(T)}\bm{h}_{k,j}^{(in)})^T}{\sqrt{D}}, \\
     & \bm{h}_{i,j}^{(T)}=\sum_{k}\mathrm{Softmax}(a_{i,k}^{(T)})\cdot (\bm{W}_V^{(T)}\bm{h}_{k,j}^{(in)}),
  \end{aligned}
\end{equation}
where $\bm{W}_Q^{(T)}$, $\bm{W}_K^{(T)}$, and $\bm{W}_V^{(T)}$ are learnable parameters defined on $\mathbb{R}^{D\times D}$.

According to Equation (\ref{eq:tattn}), in this step, each \cls{} token embedding will be updated as a linear combination of all tokens from the same channel, with the combination weights being the attention scores in $[a_{i,k}^{(T)}]$. Considering that the embeddings of \cls{} tokens will be used to predict the class label in the end, the model will learn to aggregate important information into the \cls{} tokens, which naturally makes the attention scores of the \cls{} tokens a good indicator of the importance of observation tokens \cite{qiang2022attcat}. Therefore, we propose to use the attention score of the \cls{} tokens with respect to the observation tokens to identify the \emph{pivotal tokens}. We compare such attention scores for all channels within each timepoint, and define the token with the largest attention score as the pivotal token for that timepoint.

The pivotal tokens have been well mixed within their original channels by the Temporal Attention module. However, the lack of synchronized observations along the channel dimension makes it hard for them to be mixed with tokens in other channels. Hence, we propose to copy the pivotal tokens to fill the missing tokens in other channels of the same timepoint, such that the pivotal tokens can be mixed to other channels \emph{even without synchronized tokens} with another temporal attention (Figure \ref{fig:ov}(f)). We conduct the copy-and-fill operation on the token embeddings $[\bm{h}_{i,j}^{(T)}]$, and denote the output as $[\bm{r}_{i,j}]$. Taking the $j$-th channel as an example, we compute their pairwise attention scores as
\begin{equation}
  \label{eq:mattn}
  a_{i,k}^{(M)}=\frac{(\bm{W}_Q^{(M)}\bm{r}_{i,j})(\bm{W}_K^{(M)}\bm{r}_{k,j})^T}{\sqrt{D}},
\end{equation}
where $\bm{W}_Q^{(M)}$, $\bm{W}_K^{(M)}$, and $\bm{W}_V^{(M)}$ are learnable parameters defined on $\mathbb{R}^{D\times D}$. To avoid the original information of less observed channels being overwhelmed by the pivotal tokens, we down-weight the pivotal tokens' attention score by the length of the IMTS as
\begin{equation}
  \label{eq:norm}
  \beta_{i,j}=\left\{
  \begin{aligned}
     & a_{i,j}^{(M)}/T, & \bm{h}_{i,j}^{(T)}\textrm{ is \nan} \\
     & a_{i,j}^{(M)},   & \textrm{otherwise}                  \\
  \end{aligned}
  \right.,
\end{equation}
and update the token embeddings as
\begin{equation}
  \label{eq:mupd}
  \bm{r}'_{i,j}=\sum_{k}\mathrm{Softmax}(\beta_{i,k})\cdot (\bm{W}_V^{(M)}\bm{r}_{k,j}).
\end{equation}
After this process, we reset the missing tokens in $[\bm{r}'_{i,j}]$ as \nan{} to keep the original sampling pattern \emph{unchanged}, and denote the output as $[\bm{h}_{i,j}^{(M)}]$. We refer to this process as the Token Mixing Attention.

The Channel-wise Attention module takes $[\bm{h}_{i,j}^{(M)}]$ as input, and compute the channel wise attention by treating tokens from each timepoint as a sequence. Taking the $i$-th timepoint as an example, we compute the pairwise attention scores of $[\bm{h}_{i,1}^{(M)},\ldots,\bm{h}_{i,C}^{(M)}]$, and update the token embeddings as
\begin{equation}
  \label{eq:cattn}
  \begin{aligned}
     & a_{j,k}^{(C)}=\frac{(\bm{W}_Q^{(C)}\bm{h}_{i,j}^{(M)})(\bm{W}_K^{(C)}\bm{h}_{i,k}^{(M)})^T}{\sqrt{D}}, \\
     & \bm{h}_{i,j}^{(out)}=\sum_{k}\mathrm{Softmax}(a_{k,j}^{(C)})\cdot (\bm{W}_V^{(C)}\bm{h}_{i,k}^{(M)}),
  \end{aligned}
\end{equation}
where $\bm{W}_Q^{(C)}$, $\bm{W}_K^{(C)}$, and $\bm{W}_V^{(C)}$ are learnable parameters defined on $\mathbb{R}^{D\times D}$. We denote the final output as $\bm{H}^{(out)}=[\bm{h}_{i,j}^{(out)}]$.

Note that for both Equations (\ref{eq:tattn}) and (\ref{eq:cattn}), the unobserved tokens (with embedding as \nan{}s) are masked out and do not participate in the computation.

\section{Illustrative Experimental Results}
\label{sec:exp}

In order to demonstrate the modeling ability of \shortname{}, we conduct extensive experiments on three well-adopted benchmark datasets for classification of IMTS. In this section, we first describe the experiment settings and implementation details of models in Section~\ref{sec:impl}. We then compare the performance of \shortname{} with state-of-the-art methods in Section~\ref{sec:clsf}. We further study the modeling abilities of \shortname{} for data with different degrees of channel-wise asynchrony in Section~\ref{sec:sp}, followed by an ablation study on our proposed key modules in Section~\ref{sec:ab}. To gain more insight into the design of \shortname{}, we compare performance of \shortname{} with different potential down-sampling methods in Section~\ref{sec:ds}, and analyze key hyperparameters of \shortname{} in \ref{sec:hp}. Lastly, we empirically study the efficiency of \shortname{} by comparing the number of model parameters and running speed with other baselines in Section~\ref{sec:eff}.

\begin{table}[tbp]
  \centering
  \small
  \caption{Statistics of benchmark datasets.}
  \begin{tabular}{c|c|c|c|c}
    \toprule
    Dataset & \#Samples & \#Channels & \#Classes (positive\%) & Sparsity \\
    \midrule
    P12     & 11988     & 36         & 2 (14.2\%)             & 88\%     \\
    P19     & 38303     & 34         & 2 (4.19\%)             & 95\%     \\
    PAM     & 5333      & 17         & 8 (N/A)                & 60\%     \\
    \bottomrule
  \end{tabular}%
  \label{tab:cdata}%
\end{table}%

\begin{table*}[tp]
  \centering
  \caption{Classification results. The best results are in \textbf{bold} and the second bests are \underline{underlined}.}
  \begin{tabular}{c|cc|cc|cccc}
    \toprule
    \multirow{2}[2]{*}{Methods} & \multicolumn{2}{c|}{P12}               & \multicolumn{2}{c|}{P19}               & \multicolumn{4}{c}{PAM}                                                                                                                                                                                                                             \\
    \cmidrule{2-9}              & AUROC                                  & AUPRC                                  & AUROC                                  & AUPRC                                  & Accuracy                               & Precision                              & Recall                                 & F1-Score                               \\
    \midrule
    GRU-D (2018)                & 81.9 \scriptsize{$\pm$2.1}             & 46.1 \scriptsize{$\pm$4.7}             & 83.9 \scriptsize{$\pm$1.7}             & 46.9 \scriptsize{$\pm$2.1}             & 83.3 \scriptsize{$\pm$1.6}             & 84.6 \scriptsize{$\pm$1.2}             & 85.2 \scriptsize{$\pm$1.6}             & 84.8 \scriptsize{$\pm$1.2}             \\
    SeFT (2020)                 & 73.9 \scriptsize{$\pm$2.5}             & 31.1 \scriptsize{$\pm$4.1}             & 81.2 \scriptsize{$\pm$2.3}             & 41.9 \scriptsize{$\pm$3.1}             & 67.1 \scriptsize{$\pm$2.2}             & 70.0 \scriptsize{$\pm$2.4}             & 68.2 \scriptsize{$\pm$1.5}             & 68.5 \scriptsize{$\pm$1.8}             \\
    mTAND (2020)                & 84.2 \scriptsize{$\pm$0.8}             & 48.2 \scriptsize{$\pm$3.4}             & 84.4 \scriptsize{$\pm$1.3}             & 50.6 \scriptsize{$\pm$2.0}             & 74.6 \scriptsize{$\pm$4.3}             & 74.3 \scriptsize{$\pm$4.0}             & 79.5 \scriptsize{$\pm$2.8}             & 76.8 \scriptsize{$\pm$3.4}             \\
    IP-Net (2018)               & 82.6 \scriptsize{$\pm$1.4}             & 47.6 \scriptsize{$\pm$3.1}             & 84.6 \scriptsize{$\pm$1.3}             & 38.1 \scriptsize{$\pm$3.7}             & 74.3 \scriptsize{$\pm$3.8}             & 75.6 \scriptsize{$\pm$2.1}             & 77.9 \scriptsize{$\pm$2.2}             & 76.6 \scriptsize{$\pm$2.8}             \\
    DGM2-O (2021)               & 84.4 \scriptsize{$\pm$1.6}             & 47.3 \scriptsize{$\pm$3.6}             & 86.7 \scriptsize{$\pm$3.4}             & 44.7 \scriptsize{$\pm$11.7}            & 82.4 \scriptsize{$\pm$2.3}             & 85.2 \scriptsize{$\pm$1.2}             & 83.9 \scriptsize{$\pm$2.3}             & 84.3 \scriptsize{$\pm$1.8}             \\
    MTGNN (2020)                & 74.4 \scriptsize{$\pm$6.7}             & 35.5 \scriptsize{$\pm$6.0}             & 81.9 \scriptsize{$\pm$6.2}             & 39.9 \scriptsize{$\pm$8.9}             & 83.4 \scriptsize{$\pm$1.9}             & 85.2 \scriptsize{$\pm$1.7}             & 86.1 \scriptsize{$\pm$1.9}             & 85.9 \scriptsize{$\pm$2.4}             \\
    Raindrop (2022)             & 82.8 \scriptsize{$\pm$1.7}             & 44.0 \scriptsize{$\pm$3.0}             & 87.0 \scriptsize{$\pm$2.3}             & 51.8 \scriptsize{$\pm$5.5}             & 88.5 \scriptsize{$\pm$1.5}             & 89.9 \scriptsize{$\pm$1.5}             & 89.9 \scriptsize{$\pm$0.6}             & 89.8 \scriptsize{$\pm$1.0}             \\
    ContiFormer (2023)          & 82.1 \scriptsize{$\pm$2.2}             & 44.8 \scriptsize{$\pm$3.5}             & 84.4 \scriptsize{$\pm$2.1}             & 50.4 \scriptsize{$\pm$4.3}             & 85.2 \scriptsize{$\pm$2.8}             & 86.8 \scriptsize{$\pm$2.7}             & 86.7 \scriptsize{$\pm$2.6}             & 86.3 \scriptsize{$\pm$2.7}             \\
    Coformer (2023)             & 85.8 \scriptsize{$\pm$1.9}             & 52.4 \scriptsize{$\pm$4.3}             & \underline{89.2 \scriptsize{$\pm$1.8}} & \underline{57.3 \scriptsize{$\pm$3.3}} & 91.2 \scriptsize{$\pm$0.6}             & 92.4 \scriptsize{$\pm$0.7}             & 93.7 \scriptsize{$\pm$0.7}             & 92.8 \scriptsize{$\pm$0.5}             \\
    Warpformer (2023)           & 86.5 \scriptsize{$\pm$1.2}             & 54.3 \scriptsize{$\pm$2.7}             & 88.7 \scriptsize{$\pm$2.0}             & 52.4 \scriptsize{$\pm$4.5}             & 94.2 \scriptsize{$\pm$2.3}             & 95.1 \scriptsize{$\pm$2.2}             & 94.7 \scriptsize{$\pm$2.3}             & 94.9 \scriptsize{$\pm$2.2}             \\
    ViTST (2023)                & 85.1 \scriptsize{$\pm$0.8}             & 51.1 \scriptsize{$\pm$4.1}             & \underline{89.2 \scriptsize{$\pm$2.0}} & 53.1 \scriptsize{$\pm$3.4}             & 95.8 \scriptsize{$\pm$1.3}             & 96.2 \scriptsize{$\pm$1.3}             & \underline{96.1 \scriptsize{$\pm$1.1}} & \underline{96.5 \scriptsize{$\pm$1.2}} \\
    t-PatchGNN (2024)           & 84.5 \scriptsize{$\pm$0.9}             & 50.8 \scriptsize{$\pm$2.6}             & 87.0 \scriptsize{$\pm$1.4}             & 51.5 \scriptsize{$\pm$5.2}             & 93.9 \scriptsize{$\pm$1.2}             & 94.9 \scriptsize{$\pm$0.9}             & 94.8 \scriptsize{$\pm$1.3}             & 94.8 \scriptsize{$\pm$1.2}             \\
    GraFITi (2024)              & \underline{86.6 \scriptsize{$\pm$1.1}} & \underline{54.8 \scriptsize{$\pm$3.1}} & 89.1 \scriptsize{$\pm$2.6}             & 56.6 \scriptsize{$\pm$6.3}             & \underline{96.0 \scriptsize{$\pm$0.8}} & \underline{96.3 \scriptsize{$\pm$0.6}} & 96.0 \scriptsize{$\pm$0.8}             & 96.1 \scriptsize{$\pm$0.7}             \\
    \midrule
    \shortname{} (ours)         & \textbf{88.0 \scriptsize{$\pm$1.0}}    & \textbf{58.6 \scriptsize{$\pm$4.1}}    & \textbf{90.3 \scriptsize{$\pm$2.0}}    & \textbf{58.3 \scriptsize{$\pm$5.3}}    & \textbf{97.5 \scriptsize{$\pm$0.2}}    & \textbf{97.8 \scriptsize{$\pm$0.3}}    & \textbf{97.5 \scriptsize{$\pm$0.4}}    & \textbf{97.6 \scriptsize{$\pm$0.2}}    \\
    \bottomrule
  \end{tabular}%
  \label{tab:clsf}%
\end{table*}%

\subsection{Experiment Settings}
\label{sec:impl}

In this work, we experiment on three widely used real-world healthcare and human activity IMTS datasets. Among them, the P12 \cite{goldberger2000physiobank} and P19 \cite{reyna2020early} datasets contain electronic health records as IMTS, collected from patients in their hospital stays. The objectives of analyzing these two datasets are to predict in-hospital mortality and the occurrence of sepsis, respectively. The PAM dataset \cite{reiss2012introducing} contains IMTS that measure daily living activities of the subjects with motion sensors. The objective is to predict from each IMTS a categorical label that indicates one out of eight activities of the subject.

\begin{table}[tp]
  \centering
  \small
  \setlength\tabcolsep{2.6pt}
  \caption{Statistical significance test results.}
  \begin{tabular}{c|cc|cc|cccc}
    \toprule
    \multirow{2}[2]{*}{Dataset} & \multicolumn{2}{c|}{P12} & \multicolumn{2}{c|}{P19} & \multicolumn{4}{c}{PAM}                                         \\
    \cmidrule{2-9}              & AUROC                    & AUPRC                    & AUROC                   & AUPRC & Acc.  & Prec. & Rec.  & F1    \\
    \midrule
    P-value                     & 0.008                    & 0.007                    & 0.023                   & 0.026 & 0.003 & 0.000 & 0.007 & 0.000 \\
    \bottomrule
  \end{tabular}%
  \label{tab:ttest}%
\end{table}%

The P12 and P19 datasets are highly biased regarding the target labels. Therefore, we compute the area under the receiver operating characteristic curve (AUROC) and the area under the precision-recall curve (AUPRC) of the prediction result for evaluation. For the PAM dataset, we compare the accuracy, precision, recall, and F1-score of the prediction result for evaluation. We follow previous studies \cite{zhang2022graph,li2023time} to preprocess the data, and split each dataset into $5$ subsets. Each subset is further split by $80\%$, $10\%$, and $10\%$ for training, validation, and testing, respectively. We apply the same fixed split provided by \cite{zhang2022graph} for \shortname{} and all baselines. For each dataset, we experiment with the models on the $5$ subsets, and report the mean and standard deviation of their test results. Information of the processed datasets is summarized in Table \ref{tab:cdata}.

We mainly compare our proposed \shortname{} with seven state-of-the-art methods, including Raindrop \cite{zhang2022graph}, ContiFormer \cite{chen2024contiformer}, Coformer \cite{wei2023compatible}, Warpformer \cite{zhang2023warpformer}, and ViTST \cite{li2023time}, which are the best-performing methods for IMTS classification reported in recent studies. We also compare with GraFITi \cite{yalavarthi2024grafiti} and t-PatchGNN \cite{zhang2024irregular}, which are competitive methods for IMTS forecasting, by adapting them for classification tasks.

In the experiments, we implement \shortname{} and all the baseline methods with PyTorch \cite{paszke2019pytorch}. For \shortname{} we perform grid search to obtain the best combination of hyperparameters, including the hidden dimension, number of layers, dropout rate, learning rate, and the down-sampling rate. For the baseline methods, we follow the implementation provided by the original papers. All experiments are done on an experiment platform of an NVIDIA GeForce RTX 4090 with 24 GB GPU memory.

\subsection{Comparison With State-Of-The-Art}
\label{sec:clsf}

The full results of the experiment are shown in Table \ref{tab:clsf}. In this table, we also show results of older baselines reported in past papers \cite{zhang2022graph,li2023time}. \shortname{} achieves the best performance on all evaluation metrics over the three datasets. Among the baselines, GraFITi is based on a bipartite graph representation of IMTS, which is the only method among the baselines that does not rely on any imputed values. Therefore, it has a more satisfactory performance than other baselines even though it is originally designed for forecasting. However, GraFITi does not consider the channel-wise asynchrony, thus still suffering from poor channel-wise modeling. Meanwhile, despite that Warpformer considers multi-scale modeling of IMTS, its warping-based down-sampling approach does not always mitigate the channel asynchrony, thus cannot improve its channel-wise modeling ability, which leads to a suboptimal performance. t-PatchGNN avoid the irregularity of IMTS by transforming the data into uniform and aligned temporal patches. Despite its patching approach mitigates the asynchrony to
some extent, the patching is applied only once at the beginning of their model, which is not enough to deal with different degrees of asynchrony. Its suboptimal performance also indicates that multi-scale down-sampling is needed for better channel-wise modeling. Another strong baseline is ViTST, which converts IMTS data into line-graph images, and utilizes ViTs to classify these images. Despite its simplicity in directly applying the existing ViTs, we notice that its conversion from IMTS data to images can be very complicated and time-consuming. Any changes to the raw IMTS data or the image generation settings may require a re-generation of all the images, leading to its inflexibility in applications. Its performance also relies on its backbone ViT which is pretrained on extra large image datasets. It may not perform very well if we train it from scratch like other methods in our experiments.

In comparison, \shortname{} addresses the poor channel-wise modeling problem of existing methods with a carefully-designed down-sampling module to mitigate the asynchrony, and a token mixing mechanism to enhance channel-wise feature extraction, which gives \shortname{} better channel-wise modeling ability, and makes \shortname{} significantly more powerful than other baselines. Meanwhile, \shortname{} directly works on the IMTS data and does not rely on any imputed values, which makes the data processing simple, flexible, and not subjective to extra artifacts.

We conduct statistical significance tests to further justify the improvement of \shortname{} over the baselines. On the first subset of each each dataset, we run \shortname{} and the best baseline for the dataset 5 times with different random seeds, and apply independent samples t-test to examine the significance of the improvement. Table \ref{tab:ttest} shows the p-values from the tests, which represent the statistical probabilities if there were no difference between the performances of \shortname{} and the baselines. All the p-values in Table \ref{tab:ttest} are less than 0.05. which indicates that the performance gain to be highly significant.

\begin{figure*}[tbp]
  \centering
  \includegraphics[width=0.75\linewidth]{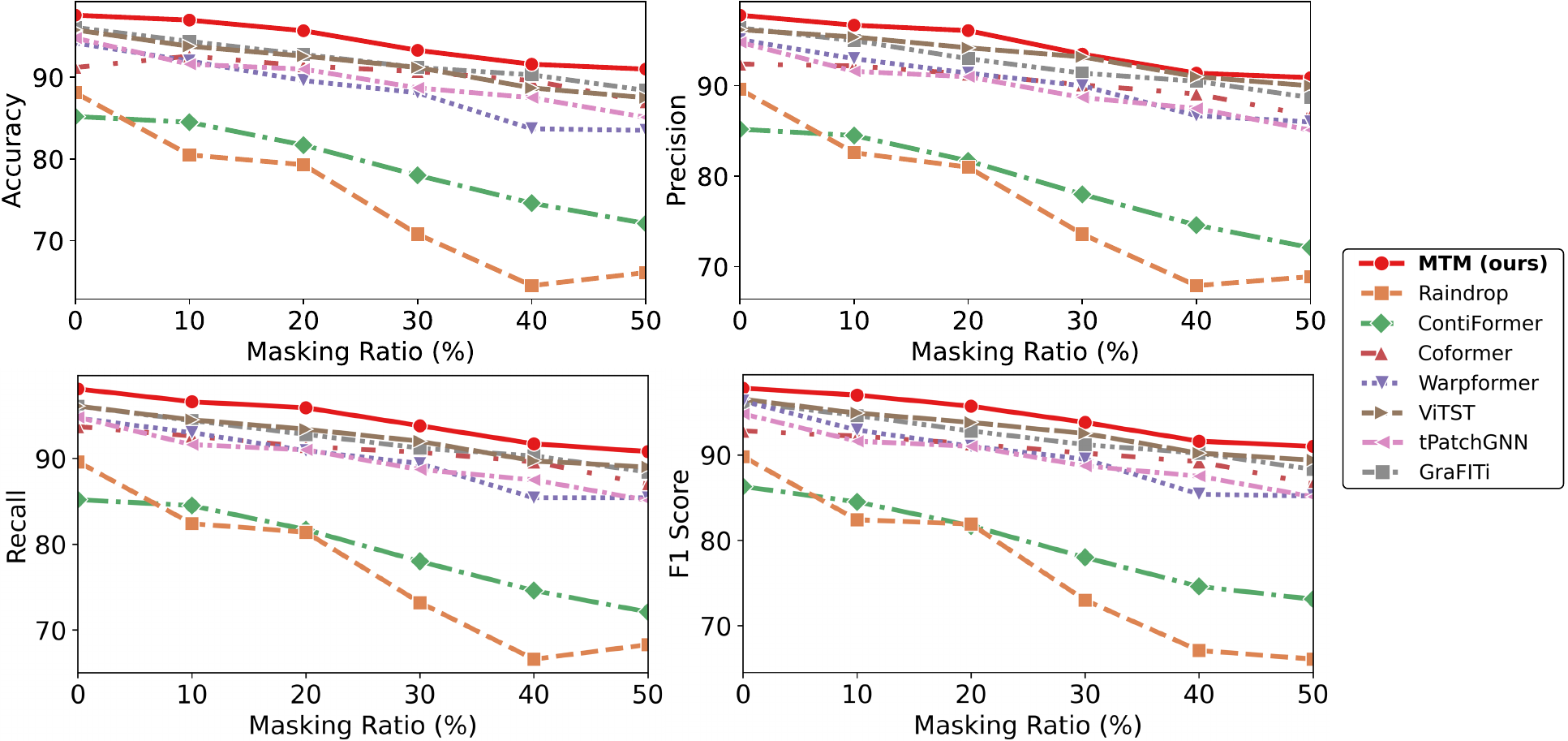}
  \caption{Performance with varying masking ratio.}
  \label{fig:miss}
\end{figure*}

\subsection{Experiments with Varying Sparsity}
\label{sec:sp}

To further study the modeling abilities of \shortname{} for data with different degrees of channel-wise asynchrony, we randomly mask out a subset of channels in the PAM dataset and compare the performance of different methods. This setting emulates real-world situations where some sensors fail or become unreachable during the data collection period. The PAM dataset is of 60\% sparsity. We experiment with 10\%-50\% data masked out, corresponding to 64\%-80\% sparsity, which emulates moderate to severe channel-wise asynchrony for our purpose. The masking is applied to the training, validation, and testing sets. We report the average F1-score over the 5 subsets as the result for each method.

As shown in Figures \ref{fig:miss}, \shortname{} leads the board with the highest F1-scores when $10\%$ to $50\%$ of the channels are masked out. With $50\%$ of the channels masked, \shortname{} can still accomplish the classification task well with an F1-score up to $91\%$. This highlights the excellent capability of \shortname{} to model IMTS data with different degrees of channel-wise asynchrony. On the other hand, we observe that the performance of Raindrop and ContiFormer flops faster than other methods when the masking ratio increases. This is because they both heavily rely on data imputation to handle the channel-wise irregularity of IMTS. Specifically, ContiFormer requires to impute missing channels in each observation to get fixed-sized vectors as its input. Raindrop learns to generate the representation of the missing channels from the observed channels. Their performance is thus more severely affected when the sparsity of the data increases.

\subsection{Ablation Study}
\label{sec:ab}

\begin{table*}[tbp]
  \centering
  \caption{Ablation study results.}
  \begin{tabular}{c|c|c|cc|cc|cccc}
    \toprule
    \multicolumn{3}{c|}{MTM Variants} & \multicolumn{2}{c|}{P12} & \multicolumn{2}{c|}{P19} & \multicolumn{4}{c}{PAM}                                                                                                                                                                                                               \\
    \midrule
    Pooling                           & \cls{}                   & Mixing                   & AUROC                      & AUPRC                      & AUROC                      & AUPRC                      & Accuracy                   & Precision                  & Recall                     & F1-Score                   \\
    \midrule
    N                                 & N                        & N                        & 86.1 \scriptsize{$\pm$1.6} & 52.6 \scriptsize{$\pm$6.6} & 87.3 \scriptsize{$\pm$3.2} & 51.1 \scriptsize{$\pm$5.5} & 95.6 \scriptsize{$\pm$1.3} & 95.9 \scriptsize{$\pm$1.3} & 95.6 \scriptsize{$\pm$1.1} & 95.8 \scriptsize{$\pm$1.2} \\
    Y                                 & N                        & N                        & 86.9 \scriptsize{$\pm$1.3} & 56.1 \scriptsize{$\pm$3.7} & 87.5 \scriptsize{$\pm$2.9} & 52.0 \scriptsize{$\pm$5.6} & 96.9 \scriptsize{$\pm$0.6} & 97.2 \scriptsize{$\pm$0.3} & 97.2 \scriptsize{$\pm$0.5} & 96.9 \scriptsize{$\pm$0.4} \\
    Y                                 & Y                        & N                        & 87.1 \scriptsize{$\pm$1.0} & 57.7 \scriptsize{$\pm$4.1} & 88.4 \scriptsize{$\pm$2.5} & 55.3 \scriptsize{$\pm$3.7} & 96.2 \scriptsize{$\pm$0.9} & 96.4 \scriptsize{$\pm$0.9} & 96.2 \scriptsize{$\pm$0.9} & 96.3 \scriptsize{$\pm$0.9} \\
    N                                 & Y                        & Y                        & 87.2 \scriptsize{$\pm$1.2} & 56.9 \scriptsize{$\pm$2.9} & 89.3 \scriptsize{$\pm$2.6} & 56.8 \scriptsize{$\pm$4.8} & 96.1 \scriptsize{$\pm$0.6} & 96.7 \scriptsize{$\pm$0.6} & 96.1 \scriptsize{$\pm$0.6} & 96.4 \scriptsize{$\pm$0.6} \\
    \midrule
    Y                                 & Y                        & Y                        & 88.0 \scriptsize{$\pm$1.0} & 58.6 \scriptsize{$\pm$4.1} & 90.3 \scriptsize{$\pm$2.0} & 58.3 \scriptsize{$\pm$5.3} & 97.5 \scriptsize{$\pm$0.2} & 97.8 \scriptsize{$\pm$0.3} & 97.5 \scriptsize{$\pm$0.4} & 97.6 \scriptsize{$\pm$0.2} \\
    \bottomrule
  \end{tabular}%
  \label{tab:abs}%
\end{table*}%

We strongly believe that the good performance of \shortname{} is rooted in our proposed multi-scale down-sampling and token mixing mechanisms. To validate the efficacy of these novel designs, we implement variants of \shortname{} by removing the Masked Concat Pooling layers, the use of \cls{} tokens, and the Token Mixing Attention modules, respectively. We carry out the experiments for these \shortname{} variants, and report the results in Table \ref{tab:abs}. All the four variants show inferior performance to the full \shortname{} model in the classification tasks. We note that by using the token mixing mechanism in \shortname{}, the AUROC increases up to 2\% on the P19 dataset. This improvement is much larger than that of the other two datasets, and also larger than the improvement brought by the down-sampling module. This is because the P19 dataset is the most sparse dataset among the three datasets. The severe lack of synchronized observations even makes the down-sampling module less helpful. On the other hand, the down-sampling modules works better than token-mixing on the PAM dataset, which is because PAM is the least sparse dataset. By comparing the variants with and without the \cls{} tokens, we discover that the introduction of the \cls{} tokens alone can also bring some performance gain to \shortname{} even without the full token mixing mechanism. These results provide strong evidence to demonstrate the effectiveness of our proposed masked concat pooling and token mixing mechanism.

\subsection{Comparing Down-Sampling Methods}
\label{sec:ds}

\begin{table*}[tbp]
  \centering
  \caption{Performance of \shortname{} with different down-sampling methods.}
  \begin{tabular}{c|cc|cc|cccc}
    \toprule
    \multirow{2}[2]{*}{Pooling Method} & \multicolumn{2}{c|}{P12}            & \multicolumn{2}{c|}{P19}            & \multicolumn{4}{c}{PAM}                                                                                                                                                                                                           \\
    \cmidrule{2-9}                     & AUROC                               & AUPRC                               & AUROC                               & AUPRC                               & Accuracy                            & Precision                           & Recall                              & F1-Score                            \\
    \midrule
    Max Pooling                        & 87.5 \scriptsize{$\pm$1.2}          & 57.3 \scriptsize{$\pm$3.8}          & 89.8 \scriptsize{$\pm$2.4}          & 56.8 \scriptsize{$\pm$3.3}          & 97.0 \scriptsize{$\pm$0.4}          & 97.4 \scriptsize{$\pm$0.5}          & 97.0 \scriptsize{$\pm$0.4}          & 97.2 \scriptsize{$\pm$0.4}          \\
    Avg Pooling                        & 87.4 \scriptsize{$\pm$1.7}          & 57.2 \scriptsize{$\pm$5.2}          & 89.5 \scriptsize{$\pm$2.2}          & 55.9 \scriptsize{$\pm$3.7}          & 96.5 \scriptsize{$\pm$0.6}          & 96.8 \scriptsize{$\pm$0.8}          & 96.5 \scriptsize{$\pm$0.6}          & 96.6 \scriptsize{$\pm$0.7}          \\
    Attention Scoring                  & 87.2 \scriptsize{$\pm$1.3}          & 56.2 \scriptsize{$\pm$3.5}          & 86.5 \scriptsize{$\pm$3.2}          & 49.9 \scriptsize{$\pm$5.4}          & 96.4 \scriptsize{$\pm$0.9}          & 96.8 \scriptsize{$\pm$0.6}          & 96.4 \scriptsize{$\pm$0.9}          & 96.6 \scriptsize{$\pm$0.7}          \\
    \textbf{Concat(max, avg)}          & \textbf{88.0 \scriptsize{$\pm$1.0}} & \textbf{58.6 \scriptsize{$\pm$4.1}} & \textbf{90.3 \scriptsize{$\pm$2.0}} & \textbf{58.3 \scriptsize{$\pm$5.3}} & \textbf{97.5 \scriptsize{$\pm$0.2}} & \textbf{97.8 \scriptsize{$\pm$0.3}} & \textbf{97.5 \scriptsize{$\pm$0.4}} & \textbf{97.6 \scriptsize{$\pm$0.2}} \\
    \bottomrule
  \end{tabular}%
  \label{tab:ds}%
\end{table*}%

\begin{figure*}[tbp]
  \centering
  \includegraphics[width=\linewidth]{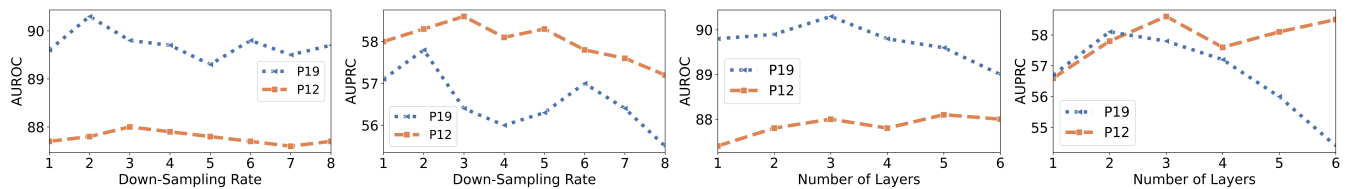}
  \caption{Performance with different hyperparameter setups.}
  \label{fig:hp}
\end{figure*}

As there is a lack of related study on the down-sampling of IMTS, we implement four representative down-sampling methods on \shortname{}. The three structured methods include max pooling, average pooling, and a concatenation of max and average pooling. The one unstructured method base on the attention score from its previous layer to select the top scored timepoints, and merge the observations to its nearest selected timepoints. We experiment these methods on our datasets, and show the results in Table \ref{tab:ds}.

From the results we observe that the masked concat pooling, which is the method we use in our \shortname{}, achieves the best performance among its structured and unstructured counterparts. We think the reason behind is that the concat pooling can maximally retain information from the data. In contrast to the regular dense data where down-sampling is applied for extracting essential information from a redundancy and achieving computational efficiency, IMTS is already very sparse, and we employ down-sampling to mitigate the channel-wise asynchrony, which makes the retaining of information more important than reducing them. This is why the concat pooling is consistently better on all the datasets despite max and average pooling are more frequently adopted for image and regular time series. On the other hand, we observe that the attention-based unstructured pooling method has inferior performance comparing the structured methods, despite its success in computer vision researches. Besides the ability of retaining information when reducing the dimension, we think another reason is that its irregular partitioning of the time dimension adds up to the difficulties for the model to learn the temporal correlations in the down-sampled timescale. We think this is also the reason why the unstructured warping-based multi-scale module in the Warpformer baseline does not actually outperform the structured pooling method we propose in \shortname{}.

\subsection{Hyperparameter Study}
\label{sec:hp}

To gain deeper insights into our proposed \shortname{} with different setups, we compare the performance of \shortname{} with different down-sampling rate and number of layers on the P12 and P19 datasets. The P12 and P19 datasets are of severe channel-wise asynchrony, which is good for us to understand the sensitivity of the related hyperparameters. The results are shown in Figure \ref{fig:hp}. When we fix the number of Token Mixing layers in \shortname{}, and experiment with different down-sampling rates, we find that the performance is suboptimal when we set the down-sampling rate too large or too small for both datasets. The best performing values are 3 for the P12 dataset and 2 for the P19 dataset. When we fix the down-sampling rates to each dataset's optimal, and experiment with varying number of layers, we find that, with increasing number of layers, the performance of \shortname{} stabilizes around the optimal value, while it flops on the P19 datasets. These findings shed light on the importance of adapting these hyperparameters to the specific properties of the datasets.

\subsection{Model Efficiency}
\label{sec:eff}

\begin{figure}[tbp]
  \centering
  \includegraphics[width=0.86\linewidth]{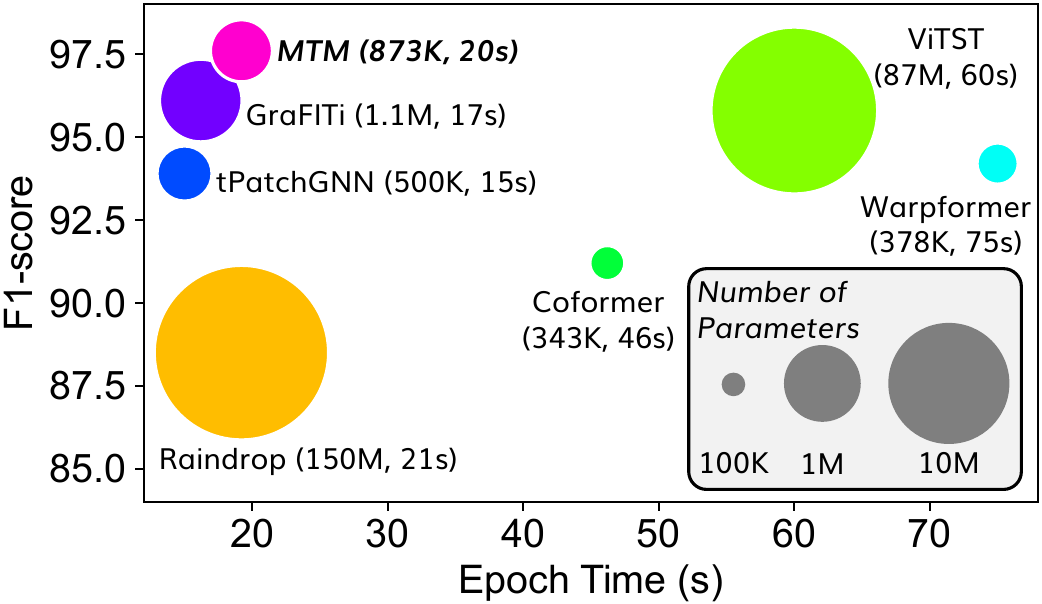}
  \caption{Model efficiency comparison.}
  \label{fig:eff}
\end{figure}

Our proposed \shortname{} mainly comprises attention operations. It does not require extra computation-intensive operations, and operations in our proposed masked concat pooling and token mixing mechanism are mostly parallelizable to enjoy accelerated computation on modern GPUs. Moreover, our multi-scale down-sampling approach by masked concat pooling allows \shortname{} to compute the attention over less tokens and further reduces the computational overhead. These characteristics enable \shortname{} to remain efficient while being very powerful. In this section, we empirically study the computation efficiency of our proposed \shortname{} by comparing the number of model parameters and training time consumption of different models on the PAM dataset. From the results shown in Figure \ref{fig:eff}, we can see that \shortname{} achieves the best performance while also maintaining a small model size and short training time. Specifically, \shortname{} runs much faster than the other two methods Warpformer and Coformer with similar Transformer-based designs. On the other hand, despite achieving a good classification performance, we can see that ViTST is significantly larger in model size and slower in training time than most methods. We also note that the Contiformer consumes significantly longer training time than all other methods because of its neural ODE-based designs. Therefore, we are not showing it in Figure \ref{fig:eff}.

\section{Conclusion}
\label{sec:conclu}

In this paper, we focus on the classification of IMTS, especially regarding the channel-wise feature extraction. We first point out the ineffectiveness of the channel-wise attention in existing Transformers to be caused by the severe channel-wise asynchrony of IMTS data, and propose to overcome this limitation in two aspects, i.e., improving the effectiveness of existing channel-wise learning module with multi-scale modeling, and introducing a new channel-wise learning mechanism. To this end, we propose \shortname{}, a multi-scale token mixing transformer. We first propose a masked concat pooling in \shortname{} to gradually down-sample IMTS. Moreover, we propose a novel channel-wise token mixing mechanism which learns to proactively choose important tokens from one channel and mix them with other channels. We experiment on three real-world datasets from the healthcare and human action recognition area, which demonstrate the applicability, feasibility, and SOTA performance of our proposed \shortname{} in real-world IMTS classification scenarios, with improvements of up to $3.8\%$ in AUPRC. Meanwhile, we also realize that the key designs of \shortname{} are currently limited to the classification task. An extension of this work to forecasting and other generative tasks of IMTS can be meaningful research directions for future efforts.

\begin{acks}
  This work was supported, in part, by \grantsponsor{rgc}{Research Grants Council Collaborative Research Fund}{https://www.ugc.edu.hk/eng/rgc/funding_opport/crf/} under grant number \grantnum{rgc}{C1045-23G}, and the \grantsponsor{uic}{Guangdong Provincial~/~Zhuhai Key Laboratory of IRADS}{https://irads.uic.edu.cn/project_application/open_funding/introduction.htm} (\grantnum{uic}{2022B1212010006}).
\end{acks}

\bibliographystyle{ACM-Reference-Format}
\balance
\bibliography{ref}

\appendix
\section{More Results on Applicability}

Our proposed \shortname{} is designed for general IMTS data. It does not rely on any domain-specific properties, which gives it great applicability to different domains. Section~\ref{sec:clsf} compares \shortname{} with baseline methods on datasets from the healthcare and human action recognition domains. To further validate the applicability of \shortname{}, we experiment on the SpokenArabicDigits (SAD) dataset \cite{bagnall2018uea} from the speech recognition domain. SAD contains 8800 regular multivariate time series of speech signals collected from native Arabic speakers' pronunciations of 10 Arabic digits. Each time series consists of 13 channels representing different feature types. The objective of this dataset is to predict the corresponding digits from the time series data. Since the data is regular, we randomly mask 80\% of the observations in each time series to generate highly sparse IMTS. We compare the classification accuracy of \shortname{} with the most competitive baselines from Table \ref{tab:clsf}. As the results shown in Table \ref{tab:sad}, our \shortname{} also achieves the best performance in comparison with these competitive baselines.

\begin{table}[htp]
  \caption{Classification accuracy on SpokenArabicDigits.}
  \begin{tabular}{c|c}
    \toprule
    Method              & Acccuracy (\%) \\
    \midrule
    Coformer            & 93.7           \\
    Warpformer          & 89.7           \\
    ViTST               & 94.0           \\
    t-PatchGNN          & 91.8           \\
    GraFITi             & 93.6           \\
    \midrule
    \textbf{MTM (ours)} & \textbf{94.5}  \\
    \bottomrule
  \end{tabular}
  \label{tab:sad}
\end{table}

\begin{table*}[htbp]
  \centering
  \caption{Classification performance on PAM with extremely high masking ratios.}
  \begin{tabular}{c|cccc|cccc|cccc|cccc}
    \toprule
    \multirow{2}[2]{*}{Methods} & \multicolumn{4}{c|}{Accuracy (\%)} & \multicolumn{4}{c|}{Precision(\%)} & \multicolumn{4}{c|}{Recall(\%)} & \multicolumn{4}{c}{F1-Score(\%)}                                                                                                                                                                                                 \\
    \cmidrule{2-17}             & 60\%                               & 70\%                               & 80\%                            & 90\%                             & 60\%          & 70\%          & 80\%          & 90\%          & 60\%          & 70\%          & 80\%          & 90\%          & 60\%          & 70\%          & 80\%          & 90\%          \\
    \midrule
    Coformer (2023)             & 85.7                               & 80.5                               & 69.7                            & 60.2                             & 84.7          & 78.6          & 69.7          & 60.2          & 85.7          & 80.5          & 68.5          & 60.2          & 85.0          & 79.2          & 68.7          & 59.5          \\
    Warpformer (2023)           & 79.4                               & 73.9                               & 63.5                            & 59.1                             & 82.5          & 77.2          & 68.3          & 62.4          & 81.5          & 76.3          & 64.0          & 59.6          & 81.4          & 76.3          & 64.9          & 59.0          \\
    ViTST (2023)                & 86.8                               & 80.5                               & 75.6                            & 69.6                             & 88.7          & 84.1          & 78.4          & 72.7          & 88.6          & 82.6          & 77.7          & \textbf{70.9} & 88.6          & 83.1          & 77.8          & 70.9          \\
    tPatchGNN (2024)            & 81.1                               & 75.5                               & 68.1                            & 59.8                             & 84.7          & 79.9          & 72.3          & 63.0          & 83.5          & 77.1          & 70.2          & 60.5          & 83.9          & 78.1          & 70.7          & 61.2          \\
    Grafiti (2024)              & 85.8                               & 80.5                               & 71.0                            & 63.2                             & 85.9          & 81.0          & 73.1          & 64.5          & 85.8          & 80.5          & 71.1          & 63.2          & 85.6          & 80.5          & 71.4          & 63.3          \\
    \midrule
    \textbf{MTM (ours)}         & \textbf{88.7}                      & \textbf{85.2}                      & \textbf{77.8}                   & \textbf{70.5}                    & \textbf{88.8} & \textbf{85.0} & \textbf{79.4} & \textbf{72.9} & \textbf{88.7} & \textbf{85.2} & \textbf{77.8} & 70.5          & \textbf{88.7} & \textbf{85.0} & \textbf{78.2} & \textbf{71.3} \\
    \bottomrule
  \end{tabular}%
  \label{tab:ext}%
\end{table*}%

\section{More Results on Extremely Sparse IMTS}

We further extend our experiments with IMTS of varying sparsity in Section~\ref{sec:sp} to extreme conditions. Table \ref{tab:ext} shows performance of \shortname{} and the most competitive baselines on the PAM dataset with 60\%-90\% data masked out, which corresponds to overall spasity of 84\%-96\%. By comparison, we observe that the advantage of \shortname{} gradually expands with the increase of data sparsity, which demonstrates the effectiveness of our proposed method in addressing the channel-wise asynchrony problem.

\section{Visualization}

\begin{figure}[htp]
  \centering
  \includegraphics[width=\linewidth]{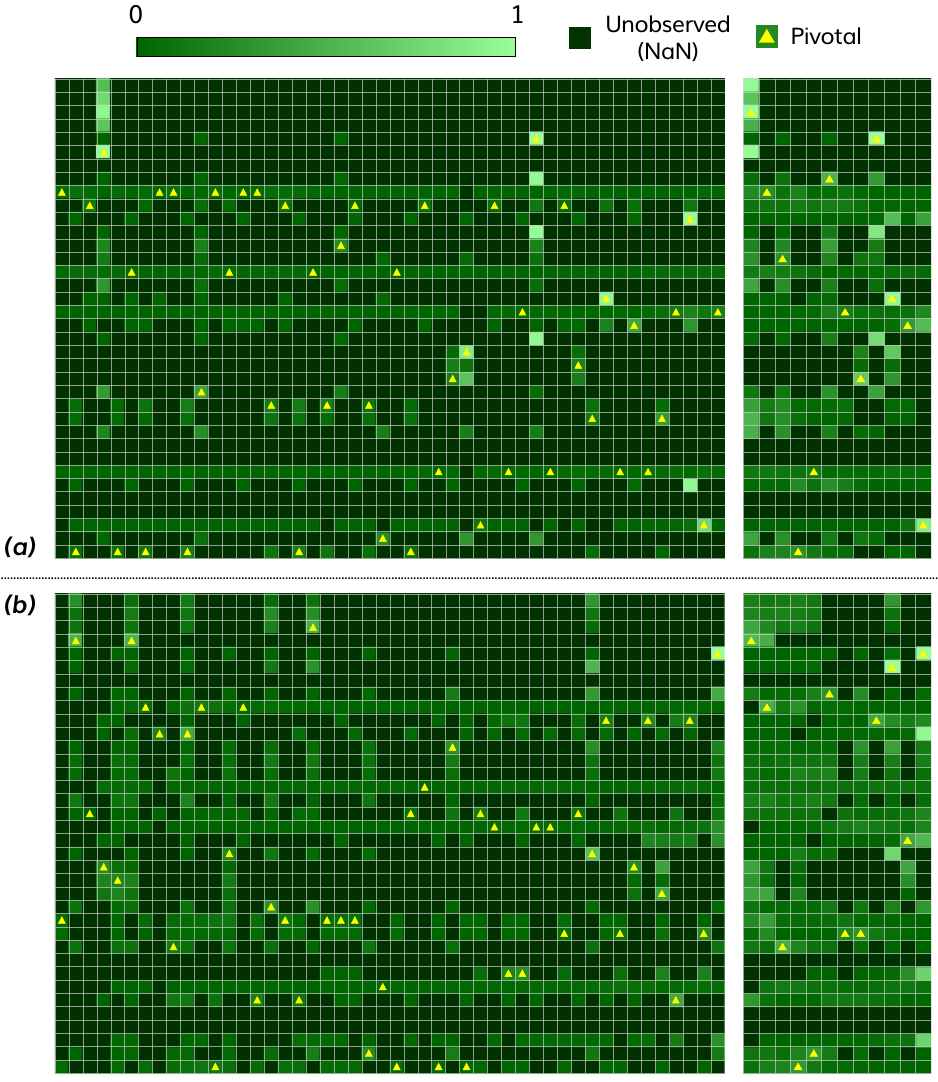}
  \caption{Attention score visualizations.}
  \label{fig:vis}
\end{figure}

To better understand the behavior of our proposed method, we study two examples of the attention scores used for choosing the pivotal tokens in our proposed token mixing mechanism. The examples are taken from the P12 dataset. For each example we visualize the attention scores $[a_{i,j}^{(T)}]$ from the first and second Token Mixing layers, where the first Token Mixing layer works on the original timescale, and the second layer works on a coarser timescale down-sampled by 4. From the attention maps in Figure \ref{fig:vis} we can see that the data at the original timescale severely suffers from channel-wise asynchrony, and a down-sampling process effectively mitigates the problem. On the other hand, tokens with the highest attention scores are chosen as the pivotal tokens and mixed with other channels in the Token Mixing Attention that follows, which further enhances \shortname{}'s channel-wise modeling ability to deal with the channel-wise asynchrony of IMTS.

\end{document}